\documentclass[conference,a4paper]{IEEEtran}
\makeatletter
\newcommand{\linebreakand}{%
  \end{@IEEEauthorhalign}
  \hfill\mbox{}\par
  \mbox{}\hfill\begin{@IEEEauthorhalign}
}
\makeatother

\IEEEoverridecommandlockouts
\usepackage{cite}
\usepackage{amsmath,amssymb,amsfonts}
\usepackage{algorithmic}
\usepackage{graphicx}
\usepackage{textcomp}
\usepackage{xcolor}
\usepackage{booktabs}
\usepackage{multirow}
\usepackage{array}
\usepackage{flafter}
\usepackage[
    a4paper,
    top=19.1mm,
    bottom=19.1mm,
    left=14.3mm,
    right=14.3mm
]{geometry}

\def\BibTeX{{\rm B\kern-.05em{\sc i\kern-.025em b}\kern-.08em
    T\kern-.1667em\lower.7ex\hbox{E}\kern-.125emX}}

\begin{document}

\flushbottom

\title{DNC-IMM: Early Lane-Change Intention Recognition via Neural Calibration Based on Driving Context Information}

\author{
\IEEEauthorblockN{1\textsuperscript{st} Woong-Chan Byun}
\IEEEauthorblockA{\textit{CCS Graduate School of Mobility} \\
\textit{Korea Advanced Institute of Science and Technology (KAIST)}\\
Daejeon, Republic of Korea, 34051\\
woongchan.byun@kaist.ac.kr}
\and
\IEEEauthorblockN{2\textsuperscript{nd} Seung-Hyun Kong\textsuperscript{*}\thanks{*Corresponding author}}
\IEEEauthorblockA{\textit{CCS Graduate School of Mobility} \\
\textit{Korea Advanced Institute of Science and Technology (KAIST)}\\
Daejeon, Republic of Korea, 34051\\
skong@kaist.ac.kr}
}

\maketitle

\begin{abstract}
Early recognition of lane-change intention is essential for proactive decision-making in autonomous driving and advanced driver assistance systems. This paper proposes a Dual Neural-Calibrated Interacting Multiple Model (DNC-IMM) that improves adaptability to driving context while preserving the probabilistic structure and interpretability of a conventional IMM. The proposed method encodes driving-context information---including target-vehicle motion, gaps to surrounding vehicles, and relative velocities---with a neural network that calibrates both the transition-probability matrix and measurement likelihoods. The final intention is determined from the calibrated IMM mode posterior rather than from a separate direct classifier. Experiments on the highD dataset demonstrate that the proposed method reliably recognizes lane-change intentions before lane crossing and provides particularly strong performance at the earlier 2--3~s prediction horizons.
\end{abstract}


\section{Introduction}

For autonomous vehicles and advanced driver assistance systems (ADAS) to exhibit safe and natural driving behavior, they must anticipate the future motion of surrounding vehicles \cite{jung2026}. Lane changes by nearby vehicles are particularly important because they are directly associated with collision risk. Accurate and timely recognition of such maneuvers is therefore a prerequisite for collision avoidance, safe-gap maintenance, and path planning \cite{cho2019,kim2014,kong2014}. Many existing studies, however, detect a lane change only after the vehicle has crossed a lane boundary, leaving insufficient time for a proactive response. In this work, we instead address early recognition at 1, 2, and 3~s before lane crossing, classifying the target vehicle into lane keeping (LK), left lane change (LCL), or right lane change (LCR). This setting is important because it estimates a developing intention from motion changes and traffic context before the maneuver is completed \cite{morris2011}.

Methods for early lane-change intention recognition can be grouped into three broad categories. First, physics-based methods such as the Interacting Multiple Model (IMM) describe vehicle motion with explicit dynamic models and infer intention through probabilistic mode transitions and measurement likelihoods. Their estimates are interpretable, but fixed transition-probability matrices and likelihood models cannot fully reflect rapidly changing context such as lane availability or interactions with nearby vehicles. Second, adaptive IMM methods modify transition probabilities and measurement likelihoods with hand-designed rules. Although these methods incorporate some context, their correction functions and weights depend heavily on expert judgment and may not generalize across driving conditions. Third, data-driven methods use deep neural networks to classify lane-change intention directly. They can learn complex patterns from large driving datasets, but their internal reasoning is often opaque and may produce predictions that are inconsistent with vehicle dynamics or roadway constraints.

Existing approaches consequently satisfy only part of the combined need for interpretability, data-driven adaptability, and awareness of surrounding traffic. A unified framework is needed that preserves the transparent probabilistic structure of a physics-based estimator while adapting its inference to changing traffic context through learned calibration.

To meet this need, we propose the Dual Neural-Calibrated IMM (DNC-IMM). The method retains the standard IMM estimation structure and augments it with neural calibration driven by target-vehicle and surrounding-traffic features. Crucially, the network does not directly output an LK, LCL, or LCR class. Instead, it calibrates the IMM mode-transition priors and measurement likelihoods, and the final intention is selected from the resulting IMM mode posterior.

The main contributions of this work are as follows:

\begin{itemize}
    \item We introduce DNC-IMM, which calibrates both the IMM transition-probability matrix and measurement likelihoods according to driving context, mitigating the limited context sensitivity of a conventional fixed-parameter IMM.
    \item The neural network outputs only transition and likelihood corrections. Because the final decision is always derived from the IMM mode posterior, the framework retains physical consistency and interpretable probabilistic reasoning.
    \item Comparative experiments demonstrate that DNC-IMM provides strong early recognition of LK, LCL, and LCR intentions at 1--3~s before lane crossing, with the best average macro F1 at the more challenging 2--3~s horizons.
\end{itemize}

\section{Related Work}

Prior work on early lane-change intention recognition can be divided into physics-based models, rule-based adaptive models, and data-driven learning models.

\subsection{Physics-Based Intention Estimation}

Physics-based approaches explicitly describe vehicle motion and infer maneuver intention through probabilistic state transitions. IMM estimators have long been used for this purpose because they maintain several motion hypotheses and update their probabilities from measurements \cite{mazor1998}. Zhang \emph{et al.} combined a Hidden Markov Model (HMM) with a Gaussian mixture model to represent lane-changing and lane-keeping behavior and used likelihood functions to interpret the behavior of a target vehicle within surrounding traffic \cite{zhang2018}. The explicit use of transition probabilities and measurement likelihoods makes these methods comparatively easy to interpret \cite{kaempchen2004}. Their transition probabilities and likelihood distributions are nevertheless usually fixed, preventing them from fully representing time-varying factors such as lane availability and dynamic vehicle interactions \cite{toledo2009,schubert2010}.

\subsection{Rule-Based Adaptive Calibration}

Adaptive methods have been introduced to reduce the rigidity of fixed physics-based models by adjusting parameters according to driving context. A Bayesian-inference-based adaptive lane-change model showed that a prediction threshold can be updated from the current traffic environment, improving transfer across road and traffic conditions \cite{wang2021}. Rule-based models such as MOBIL also express lane-change incentives and safety using physically meaningful interaction criteria \cite{kesting2007}. These approaches provide useful context sensitivity, but their rules, correction functions, and weights rely on designer assumptions. As a result, performance may degrade when the operating conditions or data distribution differ from those considered during design.

\subsection{Data-Driven Learning Models}

Recent work has increasingly used deep neural networks to predict lane-change intentions or future trajectories directly \cite{berndt2008,park2018}. Liu \emph{et al.} proposed a time-sequenced-weights HMM that assigns more importance to recent observations and reported improved intention-recognition performance over conventional HMM and recurrent models \cite{liu2023}. Maneuver-aware LSTMs and deep convolutional models have likewise demonstrated the value of learning multimodal trajectory patterns and interactions from data \cite{deo2018,cui2019}. Khelfa \emph{et al.} benchmarked MOBIL against machine-learning and ensemble techniques and found that data-driven algorithms can substantially outperform a rule-based model when trained on representative trajectories \cite{khelfa2023}. Multi-task and attention-based models have further improved long-horizon lane-change and trajectory prediction \cite{mozaffari2022,lin2021,messaoud2021}. Although these methods adapt well to nonlinear patterns, their inference process is generally a black box, making it difficult to explain a prediction or rule out a physically infeasible maneuver. DNC-IMM addresses this tradeoff by learning how to calibrate an interpretable estimator rather than replacing it with a direct classifier.

\section{Dual Neural-Calibrated IMM}
\label{sec:method}

The proposed DNC-IMM calibrates an IMM transition-probability matrix and its measurement likelihoods with neural signals while retaining the standard probabilistic estimation structure. The task is to recognize LK, LCL, or LCR at 1, 2, and 3~s before the target vehicle crosses a lane boundary. The framework consists of (i) a Neural Calibration Branch that learns corrections from target-vehicle and surrounding-traffic context and (ii) an IMM Estimation Branch that incorporates those corrections into Kalman-filter-based multi-model estimation.

\subsection{Neural Calibration Branch}
\label{subsec:neural_branch}

\begin{figure}[!htbp]
    \centering
    \includegraphics[width=\columnwidth]{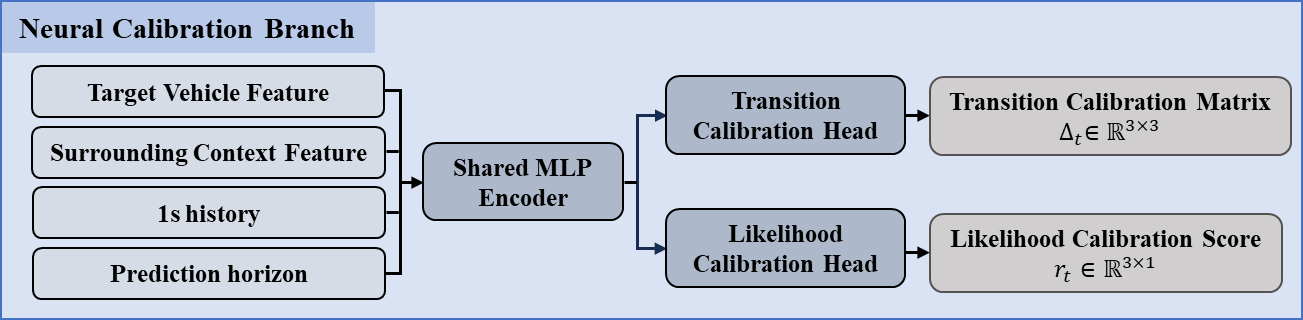}
    \caption{Architecture of the Neural Calibration Branch. A shared MLP encodes target-vehicle features, surrounding context, 1-s history statistics, and the prediction horizon. Independent heads output the transition-calibration matrix and likelihood-calibration scores.}
    \label{fig:neural_calibration}
\end{figure}

The Neural Calibration Branch compresses the context input with a shared multilayer perceptron (MLP) encoder. Two output heads then produce a transition-calibration matrix $\boldsymbol{\Delta}_t$ and a mode-wise likelihood-calibration score $\mathbf{r}_t$, respectively. Both signals are used inside the IMM Estimation Branch in Section~\ref{subsec:imm_branch}. The complete structure is shown in Fig.~\ref{fig:neural_calibration}.

\subsubsection{Input Composition and Shared Encoder}

The input to the Neural Calibration Branch is the 186-dimensional vector $\mathbf{g}_t$ summarized in Table~\ref{tab:input_features}. For both target-vehicle and surrounding-context features, statistics are computed over a 1-s history window and concatenated with a prediction-horizon indicator. Every feature is normalized with the mean and standard deviation computed from the training data, and extreme normalized values are clipped to improve inference stability.

\begin{table}[!htbp]
    \centering
    \caption{Input feature configuration of the Neural Calibration Branch.}
    \label{tab:input_features}
    \renewcommand{\arraystretch}{1.15}
    \begin{tabular}{p{0.30\columnwidth}p{0.62\columnwidth}}
        \toprule
        \textbf{Feature group} & \textbf{Key features} \\
        \midrule
        Target vehicle & Longitudinal/lateral position, velocity, acceleration, and lane geometry \\
        Surrounding context & Vehicle gaps, relative speed, distance headway, time headway, and time to collision \\
        1-s history & Mean, standard deviation, and temporal gradient over 1~s \\
        Prediction horizon & Indicator for 1--3~s early lane-change intention prediction \\
        \bottomrule
    \end{tabular}
\end{table}

The shared MLP converts $\mathbf{g}_t$ into the 64-dimensional hidden representation $\mathbf{h}_t$:

\begin{equation}
    \mathbf{h}_t = \operatorname{MLP}(\mathbf{g}_t), \qquad
    \mathbf{g}_t \in \mathbb{R}^{186},\ 
    \mathbf{h}_t \in \mathbb{R}^{64}.
    \label{eq:encoder}
\end{equation}

This operation integrates spatial and temporal cues related to lane-change intention into a compact representation. The two calibration heads share $\mathbf{h}_t$ as input but use independent parameters to generate their respective correction signals.

\subsubsection{Transition and Likelihood Calibration}

The transition-calibration head maps $\mathbf{h}_t$ to a $3\times3$ matrix. Its elements are bounded with a hyperbolic tangent:

\begin{equation}
    \boldsymbol{\Delta}_t = 3\tanh\!\left(\operatorname{FC}_{\mathrm{tr}}(\mathbf{h}_t)\right),
    \qquad \boldsymbol{\Delta}_t \in \mathbb{R}^{3\times3}.
    \label{eq:transition_head}
\end{equation}

The likelihood-calibration head similarly produces a $3\times1$ vector corresponding to LK, LCL, and LCR:

\begin{equation}
    \mathbf{r}_t = 3\tanh\!\left(\operatorname{FC}_{\mathrm{like}}(\mathbf{h}_t)\right),
    \qquad \mathbf{r}_t \in \mathbb{R}^{3\times1}.
    \label{eq:likelihood_head}
\end{equation}

In both heads, multiplication by 3 scales the $[-1,1]$ output of $\tanh$ to $[-3,3]$, thereby preventing unbounded corrections. A constant shared by all three likelihood scores cancels during the subsequent softmax operation. We therefore remove the mean and use the relative score

\begin{equation}
    r'_{t,j} = r_{t,j}
    - \frac{1}{3}\sum_{k\in\{\mathrm{LK},\mathrm{LCL},\mathrm{LCR}\}} r_{t,k},
    \quad j\in\{\mathrm{LK},\mathrm{LCL},\mathrm{LCR}\}.
    \label{eq:center_score}
\end{equation}

The outputs $\boldsymbol{\Delta}_t$ and $\mathbf{r}'_t$ are not direct maneuver predictions. They act only as calibration signals for transition priors and measurement likelihoods inside the IMM.

\subsection{IMM Estimation Branch}
\label{subsec:imm_branch}

\begin{figure}[!htbp]
    \centering
    \includegraphics[width=\columnwidth]{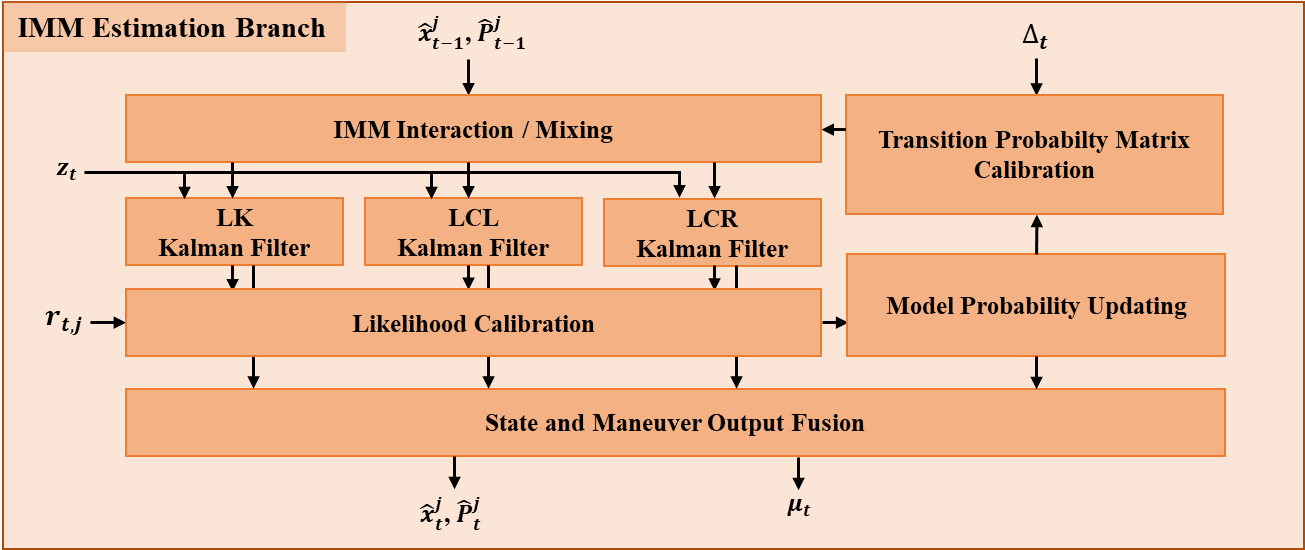}
    \caption{Architecture of the IMM Estimation Branch. The learned transition and likelihood corrections enter separate stages of the standard IMM procedure; the final state and maneuver outputs are obtained from the calibrated mode posterior.}
    \label{fig:imm_estimation}
\end{figure}

The IMM Estimation Branch receives the previous state estimate, covariance, and mode probability for each maneuver model. It performs standard IMM operations and incorporates $\boldsymbol{\Delta}_t$ and $\mathbf{r}'_t$ into the transition-probability and likelihood updates. The resulting mode posterior $\boldsymbol{\mu}_t$ provides the final maneuver decision. Figure~\ref{fig:imm_estimation} summarizes this process.

\subsubsection{State Definition and Mode-Wise Kalman Estimation}

To represent longitudinal and lateral motion jointly, the state and observation vectors are defined as

\begin{equation}
    \mathbf{x}_t =
    \begin{bmatrix}
        s_t & d_t & v_{s,t} & v_{d,t} & a_{s,t} & a_{d,t}
    \end{bmatrix}^{\!\top},
    \label{eq:state}
\end{equation}

\begin{equation}
    \mathbf{z}_t =
    \begin{bmatrix}
        s_t & d_t & v_{s,t} & v_{d,t}
    \end{bmatrix}^{\!\top}.
    \label{eq:observation}
\end{equation}

Here, $s_t$ and $d_t$ are longitudinal and lateral positions; $v_{s,t}$ and $v_{d,t}$ are the corresponding velocities; and $a_{s,t}$ and $a_{d,t}$ are longitudinal and lateral accelerations. The observation contains the measurable position and velocity terms, while acceleration is inferred during Kalman estimation.

The IMM maintains LK, LCL, and LCR simultaneously. For each mode $j$, the estimate at the previous time step is represented by

\begin{equation}
    \left\{\hat{\mathbf{x}}_{t-1}^{j},\mathbf{P}_{t-1}^{j},\mu_{t-1}^{j}\right\},
    \quad j\in\{\mathrm{LK},\mathrm{LCL},\mathrm{LCR}\},
    \label{eq:mode_state}
\end{equation}

where $\hat{\mathbf{x}}_{t-1}^{j}$ is the state estimate, $\mathbf{P}_{t-1}^{j}$ is its covariance, and $\mu_{t-1}^{j}$ is the probability that mode $j$ explains the current driving situation. Given $\mathbf{z}_t$, a Kalman update is performed independently for every mode:

\begin{equation}
    \left(\hat{\mathbf{x}}_{t}^{j},\mathbf{P}_{t}^{j}\right)
    = \operatorname{Kalman}_{j}\!\left(
    \hat{\mathbf{x}}_{t-1}^{j},\mathbf{P}_{t-1}^{j},\mathbf{z}_t\right).
    \label{eq:kalman}
\end{equation}

Thus, even under the same observation, the three filters retain distinct state and covariance estimates according to their maneuver assumptions. The IMM can consequently evaluate lane keeping and both lane-change directions in parallel rather than relying on a single motion model.

\subsubsection{Transition and Likelihood Matrix Calibration}

The transition-calibration matrix from the neural branch is combined with the base transition matrix $\boldsymbol{\Pi}_{\mathrm{base}}$ in log space:

\begin{equation}
    \boldsymbol{\Pi}_t =
    \operatorname{softmax}_{\mathrm{row}}\!\left(
    \log \boldsymbol{\Pi}_{\mathrm{base}}
    + \beta_{\mathrm{trans}}\boldsymbol{\Delta}_t\right).
    \label{eq:transition_calibration}
\end{equation}

The coefficient $\beta_{\mathrm{trans}}$ determines the strength of transition calibration. Row-wise softmax normalization ensures that every row sums to one, so $\boldsymbol{\Pi}_t$ remains a valid probability matrix. The mode prior before the current measurement is then

\begin{equation}
    \bar{\boldsymbol{\mu}}_t
    = \boldsymbol{\Pi}_t\boldsymbol{\mu}_{t-1}.
    \label{eq:mode_prior}
\end{equation}

Next, the centered likelihood score calibrates the measurement likelihood calculated by each Kalman filter. If $\log\Lambda_t^j$ denotes the original log likelihood of mode $j$, its calibrated value is

\begin{equation}
    \log\widetilde{\Lambda}_t^j = \log\Lambda_t^j
    + \operatorname{clip}\!\left(
    \alpha_{\mathrm{like}}r'_{t,j},-2,2\right).
    \label{eq:likelihood_calibration}
\end{equation}

The coefficient $\alpha_{\mathrm{like}}$ controls the calibration strength, while clipping prevents an excessive likelihood correction. The final mode posterior is obtained by combining the calibrated prior and likelihood:

\begin{equation}
    \boldsymbol{\mu}_t =
    \operatorname{softmax}\!\left(
    \log\bar{\boldsymbol{\mu}}_t
    + \log\widetilde{\boldsymbol{\Lambda}}_t\right).
    \label{eq:posterior}
\end{equation}

The final maneuver estimate is selected directly from this posterior,

\begin{equation}
    \hat{y}_t = \underset{j\in\{\mathrm{LK},\mathrm{LCL},\mathrm{LCR}\}}{\arg\max}
    \ \mu_t^j.
    \label{eq:decision}
\end{equation}

Accordingly, the network cannot bypass the IMM with a direct classification output; every decision remains traceable to the calibrated mode probabilities.

\subsubsection{Posterior-Alignment Loss}

The final IMM posterior in \eqref{eq:posterior}, rather than a separate classifier output, is used as the learning target. Training encourages the transition and likelihood corrections to increase the posterior probability of the ground-truth maneuver. The total loss is

\begin{equation}
    \mathcal{L} = \mathcal{L}_{\mathrm{post}}
    + \lambda_{\Delta}\mathcal{L}_{\Delta}
    + \lambda_{\mathrm{invalid}}\mathcal{L}_{\mathrm{invalid}}.
    \label{eq:loss}
\end{equation}

Here, $\mathcal{L}_{\mathrm{post}}$ aligns the posterior with the ground-truth mode. The regularizer $\mathcal{L}_{\Delta}$ limits the magnitude of $\boldsymbol{\Delta}_t$, with $\lambda_{\Delta}$ controlling its contribution. The lane-availability term $\mathcal{L}_{\mathrm{invalid}}$ suppresses LCL or LCR when the corresponding adjacent lane does not exist, and $\lambda_{\mathrm{invalid}}$ determines the strength of this constraint. The objective therefore combines posterior alignment with safeguards against overly large neural corrections and physically impossible lane-change predictions.

\section{Experimental Results}
\label{sec:experiments}

We evaluate DNC-IMM on the highD dataset and measure the accuracy of LK, LCL, and LCR mode posteriors at 1, 2, and 3~s before lane crossing. The experiments also isolate the contributions of transition and likelihood calibration.

\subsection{Experimental Setup}

The highD dataset contains vehicle trajectories recorded by drones over German highways \cite{krajewski2018}. It provides time-indexed position, velocity, acceleration, lane ID, vehicle size, and recording-specific road information for each vehicle. These data allow us to compute relative gaps and velocities, lane availability, and lane-crossing times. The model input includes target-vehicle kinematics, lane information, relative gaps and velocities to surrounding vehicles, and short-history statistics. All features are normalized using statistics from the training split.

To prevent trajectories from the same recording from appearing in both training and evaluation, the data are split by recording: 43 recordings for training, 5 for validation, and 12 for testing. Predictions are evaluated for LK, LCL, and LCR at 1, 2, and 3~s before lane crossing. For every input, the network generates transition and likelihood corrections, which are incorporated into the IMM before calculating the final mode posterior.

We use the AdamW optimizer with a learning rate of $0.001$, weight decay of $0.0001$, batch size of 1024, and 20 training epochs. The calibration coefficients are $\alpha_{\mathrm{like}}=1.0$ and $\beta_{\mathrm{trans}}=1.5$, while the loss weights are $\lambda_{\Delta}=0.0001$ and $\lambda_{\mathrm{invalid}}=0.5$. All experiments are performed on an NVIDIA GeForce RTX~3070 GPU.

\subsection{Comparative Results}

Performance is evaluated with class-wise F1 and macro F1. For mode $j$, class-wise F1 is the harmonic mean of precision and recall:

\begin{equation}
    \mathrm{F1}_{j} =
    \frac{2\,\mathrm{Precision}_{j}\,\mathrm{Recall}_{j}}
    {\mathrm{Precision}_{j}+\mathrm{Recall}_{j}},
    \quad j\in\{\mathrm{LK},\mathrm{LCL},\mathrm{LCR}\}.
    \label{eq:f1}
\end{equation}

Macro F1 gives equal weight to the three classes and is therefore suitable for the class imbalance in early lane-change recognition:

\begin{equation}
    \mathrm{Macro\ F1} = \frac{1}{3}
    \sum_{j\in\{\mathrm{LK},\mathrm{LCL},\mathrm{LCR}\}}\mathrm{F1}_{j}.
    \label{eq:macro_f1}
\end{equation}

Table~\ref{tab:comparison} compares DNC-IMM with prior methods at each prediction horizon. DNC-IMM is slightly below the physics-informed method of Shi \emph{et al.} at 1~s and, consequently, in the average over all three horizons \cite{shi2026}. At the earlier and more challenging 2--3~s horizons, however, DNC-IMM obtains the highest average macro F1 of 0.9185. It also exceeds the long-horizon averages of the multi-task attention model of Mozaffari \emph{et al.} \cite{mozaffari2022} and the time-sequenced-weights HMM of Liu \emph{et al.} \cite{liu2023}. These results indicate that neural calibration of an IMM is especially effective before a lane-change maneuver becomes visually or kinematically obvious.

\begin{table}[!htbp]
    \centering
    \caption{Comparative macro F1 performance on the test set. Higher values are better.}
    \label{tab:comparison}
    \scriptsize
    \renewcommand{\arraystretch}{1.08}
    \setlength{\tabcolsep}{2.2pt}
    \begin{tabular*}{\columnwidth}{@{\extracolsep{\fill}}lccccc@{}}
        \toprule
        \textbf{Method} & \textbf{1 s} & \textbf{2 s} & \textbf{3 s} & \textbf{2--3 s} & \textbf{Mean} \\
        \midrule
        Shi \emph{et al.} \cite{shi2026} & \textbf{0.9802} & 0.9536 & 0.8789 & 0.9162 & \textbf{0.9376} \\
        Mozaffari \emph{et al.} \cite{mozaffari2022} & 0.9750 & 0.9494 & 0.8629 & 0.9061 & 0.9291 \\
        Liu \emph{et al.} \cite{liu2023} & 0.7283 & 0.7743 & 0.5977 & 0.6860 & 0.7001 \\
        \textbf{DNC-IMM (ours)} & 0.9726 & \textbf{0.9567} & \textbf{0.8804} & \textbf{0.9185} & 0.9366 \\
        \bottomrule
    \end{tabular*}
\end{table}

\begin{figure}[!htbp]
    \centering
    \includegraphics[width=\columnwidth]{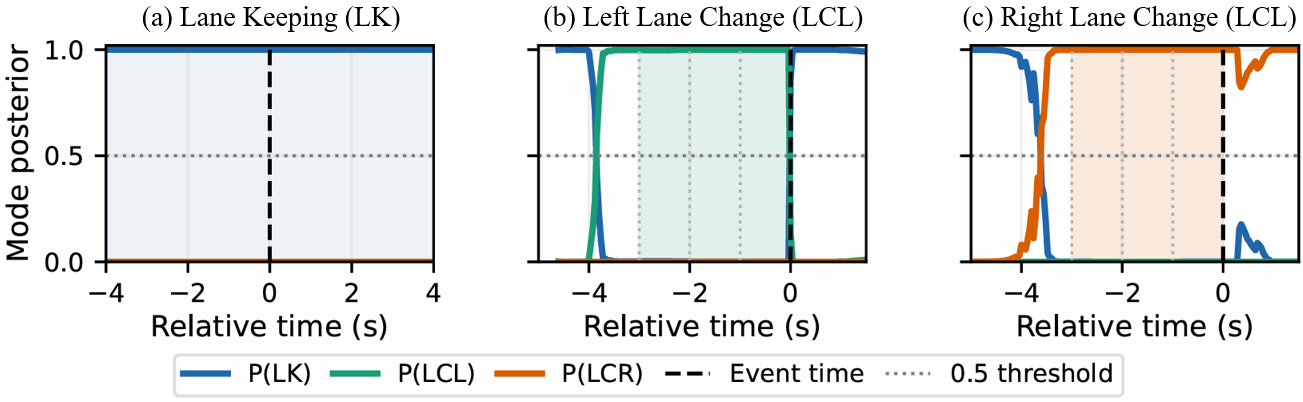}
    \caption{IMM mode probabilities for representative LK, LCL, and LCR scenarios from highD recording~59. The dashed vertical line marks the lane-crossing event and the dotted horizontal line denotes the 0.5 decision threshold.}
    \label{fig:mode_probabilities}
\end{figure}

Figure~\ref{fig:mode_probabilities} visualizes the temporal mode probabilities for representative LK, LCL, and LCR scenarios. In the LK scenario, $P(\mathrm{LK})$ remains high and the two lane-change probabilities remain low, showing that the model suppresses unnecessary lane-change predictions. In the LCL and LCR scenarios, $P(\mathrm{LK})$ decreases before lane crossing while the posterior for the true lane-change direction increases. In both cases, the corresponding probability exceeds 0.5 before the vehicle crosses the lane boundary. Thus, DNC-IMM recognizes intention during the developing maneuver rather than only after the lane change has completed.

Together, the quantitative results and posterior trajectories show that DNC-IMM provides not only a final class label but also an interpretable account of how the probability mass moves from LK toward LCL or LCR as an intention develops.

\subsection{Component Ablation}

We compare four configurations to measure the contribution of each neural correction. DNC-IMM uses both corrections; Transition-only and Likelihood-only each retain a single correction; and No Neural IMM removes both learned calibrations.

\begin{table}[!htbp]
    \centering
    \begin{minipage}{\columnwidth}
    \centering
    \caption{Component ablation results on the test set.}
    \label{tab:ablation}
    \renewcommand{\arraystretch}{1.12}
    \resizebox{\linewidth}{!}{%
    \begin{tabular}{lcccc}
        \toprule
        \textbf{Method} & \textbf{F1@1s $\uparrow$} & \textbf{F1@2s $\uparrow$} & \textbf{F1@3s $\uparrow$} & \textbf{Mean Macro F1 $\uparrow$} \\
        \midrule
        Likelihood-only IMM & 0.9600 & 0.9500 & 0.8767 & 0.9289 \\
        Transition-only IMM & 0.9578 & 0.9395 & 0.8426 & 0.9133 \\
        No Neural IMM & 0.5335 & 0.5430 & 0.4487 & 0.5084 \\
        \textbf{DNC-IMM (proposed)} & \textbf{0.9726} & \textbf{0.9567} & \textbf{0.8804} & \textbf{0.9366} \\
        \bottomrule
    \end{tabular}
    }
    \end{minipage}
\end{table}

As shown in Table~\ref{tab:ablation}, DNC-IMM achieves the highest macro F1 at every horizon and the best overall mean of 0.9366. Likelihood-only reaches 0.9289, outperforming Transition-only at 0.9133, which indicates that context-aware measurement-likelihood calibration makes a particularly strong contribution. No Neural IMM yields only 0.5084 and tends toward conservative LK predictions. Both learned corrections are therefore necessary for the strongest and most balanced early recognition.

\begin{figure}[!htbp]
    \centering
    \includegraphics[width=\columnwidth]{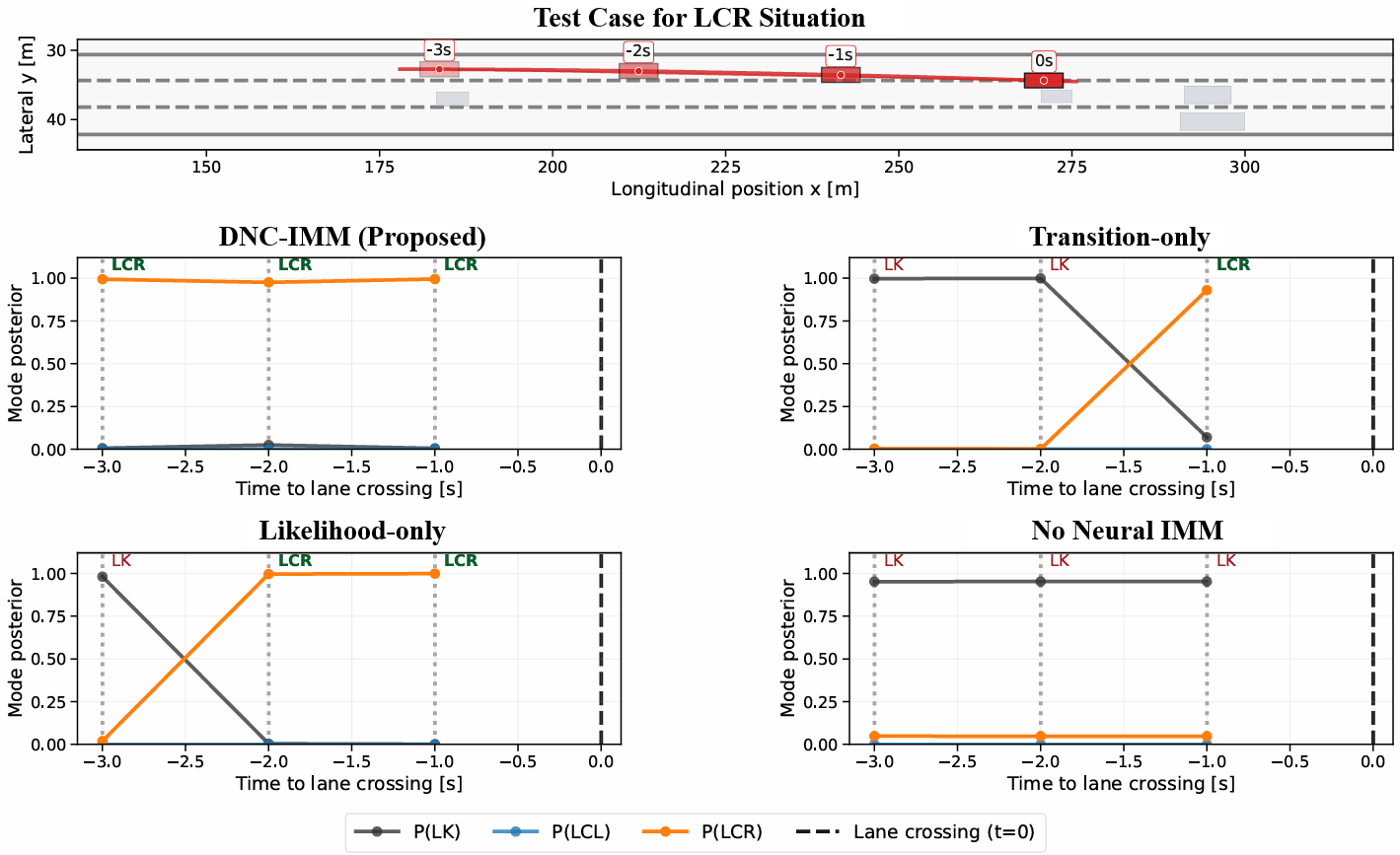}
    \caption{Qualitative component ablation for a representative LCR test case. DNC-IMM identifies LCR from 3~s before lane crossing, whereas single-calibration variants respond later and the uncalibrated IMM remains biased toward LK.}
    \label{fig:component_ablation}
\end{figure}

Figure~\ref{fig:component_ablation} compares the posterior trajectories for a representative LCR test case. DNC-IMM correctly predicts LCR at every evaluated horizon from 3~s onward and maintains a high $P(\mathrm{LCR})$. Transition-only switches to LCR only at 1~s, and Likelihood-only begins to predict LCR at 2~s but still predicts LK at 3~s. No Neural IMM predicts LK at every horizon. The case confirms that combining both corrections enables the earliest and most stable recognition.

\subsection{Case Analysis of Neural Calibration Signals}

\begin{figure}[!htbp]
    \centering
    \includegraphics[width=\columnwidth]{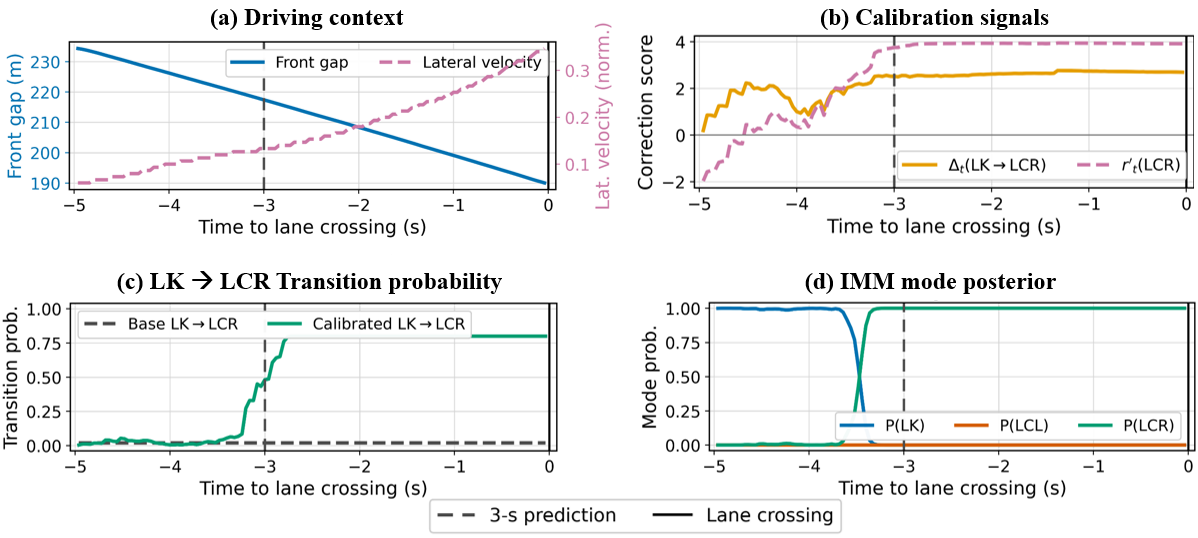}
    \caption{Neural calibration and mode-posterior changes in an LCR case. Driving context is transformed into transition and likelihood corrections, which increase the calibrated LK-to-LCR transition probability and ultimately move the posterior from LK to LCR before lane crossing.}
    \label{fig:calibration_case}
\end{figure}

Figure~\ref{fig:calibration_case} aligns the major variables in an LCR case along the same time axis. Panel (a) shows the front gap in the adjacent lane and lateral velocity in the lane-change direction. Panel (b) shows the LK-to-LCR transition correction $\Delta_t(\mathrm{LK}\!\rightarrow\!\mathrm{LCR})$ and the LCR likelihood correction $r'_t(\mathrm{LCR})$. Panel (c) compares the base and calibrated LK-to-LCR transition probabilities, and panel (d) presents the final IMM mode posterior.

The case illustrates the complete reasoning chain. Before lane crossing, lateral motion and the adjacent-lane front gap change together in panel (a). The neural branch responds by increasing both the LCR likelihood correction and the LK-to-LCR transition correction in panel (b). Consequently, the calibrated transition probability in panel (c) rises substantially above the base value. The final posterior in panel (d) then shifts from LK to LCR before the maneuver crosses the lane boundary. DNC-IMM therefore connects observed driving context to explicit correction signals, a calibrated transition matrix, calibrated measurement likelihoods, and an interpretable final posterior.

\section{Conclusion}

This paper proposed DNC-IMM for early lane-change intention recognition under surrounding-vehicle context. The method preserves the probabilistic structure of a conventional IMM while incorporating neural calibration of both transition probabilities and measurement likelihoods. Rather than using a separate direct classifier, it determines LK, LCL, or LCR from the calibrated IMM mode posterior. Experiments on highD demonstrate stable recognition before lane crossing, with particularly strong performance in the important 2--3~s early-prediction range. Component ablations further show that using transition and likelihood calibration together outperforms either correction alone and greatly improves upon an uncalibrated IMM. Finally, the calibration-signal case study demonstrates how driving context propagates through interpretable corrections to the final posterior. These results show that DNC-IMM can incorporate data-driven context sensitivity while retaining physically grounded, probabilistic reasoning.

\section*{Acknowledgment}

This work was supported by the Ministry of Trade, Industry and Resources (MOTIR), Republic of Korea, under Grant RS-2025-25451359.

\bibliographystyle{IEEEtran}
\bibliography{DNC_IMM}

@inproceedings{berndt2008,
  author    = {Holger Berndt and Jorg Emmert and Klaus Dietmayer},
  title     = {Continuous Driver Intention Recognition with Hidden Markov Models},
  booktitle = {2008 11th International IEEE Conference on Intelligent Transportation Systems},
  year      = {2008},
  pages     = {1189--1194},
  doi       = {10.1109/ITSC.2008.4732630},
}

@inproceedings{cho2019,
  author    = {Sang Jae Cho and Bo Seong Kim and Tae Seon Kim and Seung-Hyun Kong},
  title     = {Enhancing {GNSS} Performance and Detection of Road Crossing in Urban Area Using Deep Learning},
  booktitle = {2019 IEEE Intelligent Transportation Systems Conference},
  year      = {2019},
  pages     = {2115--2120},
  doi       = {10.1109/ITSC.2019.8917224},
}

@inproceedings{cui2019,
  author    = {Henggang Cui and Vladan Radosavljevic and Fang-Chieh Chou and Tsung-Han Lin and Thi Nguyen and Tzu-Kuo Huang and Jeff Schneider and Nemanja Djuric},
  title     = {Multimodal Trajectory Predictions for Autonomous Driving Using Deep Convolutional Networks},
  booktitle = {2019 International Conference on Robotics and Automation},
  year      = {2019},
  pages     = {2090--2096},
  doi       = {10.1109/ICRA.2019.8793868},
}

@inproceedings{deo2018,
  author    = {Nachiket Deo and Mohan M. Trivedi},
  title     = {Multi-Modal Trajectory Prediction of Surrounding Vehicles with Maneuver Based {LSTM}s},
  booktitle = {2018 IEEE Intelligent Vehicles Symposium},
  year      = {2018},
  pages     = {1179--1184},
  doi       = {10.1109/IVS.2018.8500493},
}

@inproceedings{jung2026,
  author    = {Hee-Yang Jung and Dong-Hee Paek and Seung-Hyun Kong},
  title     = {Open-Source Autonomous Driving Software Platforms: Comparison of {Autoware} and {Apollo}},
  booktitle = {2026 IEEE Intelligent Vehicles Symposium},
  year      = {2026},
  pages     = {1974--1981},
}

@inproceedings{kaempchen2004,
  author    = {Nico Kaempchen and Kristian Weiss and Michael Schaefer and Klaus C. J. Dietmayer},
  title     = {{IMM} Object Tracking for High Dynamic Driving Maneuvers},
  booktitle = {2004 IEEE Intelligent Vehicles Symposium},
  year      = {2004},
  pages     = {825--830},
  doi       = {10.1109/IVS.2004.1336491},
}

@article{kesting2007,
  author  = {Arne Kesting and Martin Treiber and Dirk Helbing},
  title   = {General Lane-Changing Model {MOBIL} for Car-Following Models},
  journal = {Transportation Research Record},
  year    = {2007},
  volume  = {1999},
  number  = {1},
  pages   = {86--94},
  doi     = {10.3141/1999-10},
}

@article{khelfa2023,
  author  = {Basma Khelfa and Ibrahima Ba and Antoine Tordeux},
  title   = {Predicting Highway Lane-Changing Maneuvers: A Benchmark Analysis of Machine and Ensemble Learning Algorithms},
  journal = {Physica A: Statistical Mechanics and its Applications},
  year    = {2023},
  volume  = {612},
  pages   = {128471},
  doi     = {10.1016/j.physa.2023.128471},
}

@article{kim2014,
  author  = {Binhee Kim and Seung-Hyun Kong},
  title   = {Two-Dimensional Compressed Correlator for Fast Acquisition of {BOC}($m,n$) Signals},
  journal = {IEEE Transactions on Vehicular Technology},
  year    = {2014},
  volume  = {63},
  number  = {6},
  pages   = {2662--2672},
  doi     = {10.1109/TVT.2013.2293225},
}

@article{kong2014,
  author  = {Seung-Hyun Kong},
  title   = {Fast Multi-Satellite {ML} Acquisition for {A-GPS}},
  journal = {IEEE Transactions on Wireless Communications},
  year    = {2014},
  volume  = {13},
  number  = {9},
  pages   = {4935--4946},
  doi     = {10.1109/TWC.2014.2327101},
}

@inproceedings{krajewski2018,
  author    = {Robert Krajewski and Julian Bock and Laurent Kloeker and Lutz Eckstein},
  title     = {The {highD} Dataset: A Drone Dataset of Naturalistic Vehicle Trajectories on German Highways for Validation of Highly Automated Driving Systems},
  booktitle = {2018 21st International Conference on Intelligent Transportation Systems},
  year      = {2018},
  pages     = {2118--2125},
  doi       = {10.1109/ITSC.2018.8569552},
}

@article{lin2021,
  author  = {Lei Lin and Weizi Li and Huikun Bi and Lingqiao Qin},
  title   = {Vehicle Trajectory Prediction Using {LSTM}s With Spatial--Temporal Attention Mechanisms},
  journal = {IEEE Intelligent Transportation Systems Magazine},
  year    = {2022},
  volume  = {14},
  number  = {2},
  pages   = {197--208},
  doi     = {10.1109/MITS.2021.3049404},
}

@article{liu2023,
  author  = {Pujun Liu and Ting Qu and Huihua Gao and Xun Gong},
  title   = {Driving Intention Recognition of Surrounding Vehicles Based on a Time-Sequenced Weights Hidden Markov Model for Autonomous Driving},
  journal = {Sensors},
  year    = {2023},
  volume  = {23},
  number  = {21},
  pages   = {8761},
  doi     = {10.3390/s23218761},
}

@article{mazor1998,
  author  = {Eshed Mazor and Amir Averbuch and Yaakov Bar-Shalom and Joshua Dayan},
  title   = {Interacting Multiple Model Methods in Target Tracking: A Survey},
  journal = {IEEE Transactions on Aerospace and Electronic Systems},
  year    = {1998},
  volume  = {34},
  number  = {1},
  pages   = {103--123},
  doi     = {10.1109/7.640267},
}

@article{messaoud2021,
  author  = {Kaouther Messaoud and Itheri Yahiaoui and Anne Verroust-Blondet and Fawzi Nashashibi},
  title   = {Attention Based Vehicle Trajectory Prediction},
  journal = {IEEE Transactions on Intelligent Vehicles},
  year    = {2021},
  volume  = {6},
  number  = {1},
  pages   = {175--185},
  doi     = {10.1109/TIV.2020.2991952},
}

@inproceedings{morris2011,
  author    = {Brendan Morris and Anup Doshi and Mohan M. Trivedi},
  title     = {Lane Change Intent Prediction for Driver Assistance: On-Road Design and Evaluation},
  booktitle = {2011 IEEE Intelligent Vehicles Symposium},
  year      = {2011},
  pages     = {895--901},
  doi       = {10.1109/IVS.2011.5940538},
}

@article{mozaffari2022,
  author  = {Sajjad Mozaffari and Eduardo Arnold and Mehrdad Dianati and Saber Fallah},
  title   = {Early Lane Change Prediction for Automated Driving Systems Using Multi-Task Attention-Based Convolutional Neural Networks},
  journal = {IEEE Transactions on Intelligent Vehicles},
  year    = {2022},
  volume  = {7},
  number  = {3},
  pages   = {758--770},
  doi     = {10.1109/TIV.2022.3161785},
}

@inproceedings{park2018,
  author    = {Seong Hyeon Park and ByeongDo Kim and Chang Mook Kang and Chung Choo Chung and Jun Won Choi},
  title     = {Sequence-to-Sequence Prediction of Vehicle Trajectory via {LSTM} Encoder-Decoder Architecture},
  booktitle = {2018 IEEE Intelligent Vehicles Symposium},
  year      = {2018},
  pages     = {1672--1678},
  doi       = {10.1109/IVS.2018.8500658},
}

@article{schubert2010,
  author  = {Robin Schubert and Karsten Schulze and Gerd Wanielik},
  title   = {Situation Assessment for Automatic Lane-Change Maneuvers},
  journal = {IEEE Transactions on Intelligent Transportation Systems},
  year    = {2010},
  volume  = {11},
  number  = {3},
  pages   = {607--616},
  doi     = {10.1109/TITS.2010.2049353},
}

@inproceedings{shi2026,
  author    = {Jiazhao Shi and Yichen Lin and Y. Hua and Z. Wang and Z. Zhang and W. Zheng and Y. Song and K. Lu and S. Lu},
  title     = {Multiscenario Highway Lane-Change Intention Prediction: A Physics-Informed {AI} Framework for Three-Class Classification},
  booktitle = {International Conference on Smart Transportation and City Engineering (STCE 2025)},
  series    = {Proceedings of SPIE},
  year      = {2026},
  volume    = {14120},
  pages     = {129--145},
  note      = {Art. no. 141200L},
  doi       = {10.1117/12.3101408},
}

@article{toledo2009,
  author  = {Rafael Toledo-Moreo and Miguel A. Zamora-Izquierdo},
  title   = {{IMM}-Based Lane-Change Prediction in Highways with Low-Cost {GPS/INS}},
  journal = {IEEE Transactions on Intelligent Transportation Systems},
  year    = {2009},
  volume  = {10},
  number  = {1},
  pages   = {180--185},
  doi     = {10.1109/TITS.2008.2011691},
}

@article{zhang2018,
  author  = {Yihuan Zhang and Qin Lin and Jun Wang and Sicco Verwer and John M. Dolan},
  title   = {Lane-Change Intention Estimation for Car-Following Control in Autonomous Driving},
  journal = {IEEE Transactions on Intelligent Vehicles},
  year    = {2018},
  volume  = {3},
  number  = {3},
  pages   = {276--286},
  doi     = {10.1109/TIV.2018.2843178},
}

@article{wang2021,
  author  = {Jinghua Wang and Zhao Zhang and Guangquan Lu},
  title   = {A Bayesian Inference Based Adaptive Lane Change Prediction Model},
  journal = {Transportation Research Part C: Emerging Technologies},
  year    = {2021},
  volume  = {132},
  pages   = {103363},
  doi     = {10.1016/j.trc.2021.103363},
}

\end{document}